\documentclass[runningheads]{llncs}

\usepackage{eccv}

\usepackage{eccvabbrv}

\usepackage{graphicx}
\usepackage{wrapfig}
\usepackage{subcaption}
\usepackage{booktabs}
\usepackage[ruled,vlined,linesnumbered]{algorithm2e}
\usepackage[accsupp]{axessibility}  
\usepackage{booktabs}
\usepackage{multirow}
\usepackage[table]{xcolor}

\usepackage{hyperref}

\usepackage{orcidlink}

\usepackage{makecell}
\begin{document}

\title{VLA-Hijack: A Transferable Patch Attack against Vision-Language-Action Models via Visual Proprioception Hijacking} 
\titlerunning{Abbreviated paper title}

\author{
    Jiyuan Fu$^{1,*}$, 
    Kaixun Jiang$^{1,*}$, 
    Jingkai Jia$^{1}$, 
    Zhaoyu Chen$^{1}$, 
    Xueyao Chen$^{1}$, \\
    Lingyi Hong$^{1}$, 
    Shuyong Gao$^{2}$, 
    Chenzhi Tan$^{1}$, 
    Dingkang Yang$^{1}$, \\
    Wenqiang Zhang$^{1,\dagger}$
    \vspace{0.3cm} \\
    $^{1}$ Fudan University \\
    $^{2}$ The Hong Kong Polytechnic University \\
    \vspace{0.2cm}
    \small $^*$ Equal contribution \quad $^{\dagger}$ Corresponding author
}

\authorrunning{J.~Fu, K.~Jiang, et al.}

\institute{~}

\maketitle

\begin{abstract}
  While Vision-Language-Action (VLA) models have emerged as powerful generalist policies, their severe vulnerability to adversarial patches significantly hinders their deployment in safety-critical domains. Moreover, existing patch attacks primarily focus on white-box settings, heavily overfitting to the specific action output space of the target model, which results in poor cross-architecture transferability. To overcome this limitation, we propose VLA-Hijack, a unified adversarial framework that breaks the transferability bottleneck by exploiting a fundamental vulnerability identified in this work: before planning any motion, a VLA model must first use visual information to locate its own robotic arm within the environment. Targeting this shared visual self-localization process, our approach concurrently optimizes Attention-Guided Proprioceptive Suppression to inhibit the real robotic arm's features, and Multimodal Proprioceptive Injection to establish the patch as a surrogate "phantom embodiment". By alternating between semantic concept anchoring and visual prototype projection, VLA-Hijack effectively severs the semantic relationship between the agent's true embodiment and its control policy. Extensive experiments across diverse architectures (OpenVLA, UniVLA, and CronusVLA) demonstrate that VLA-Hijack achieves superior optimization efficiency in white-box settings and sets a new SOTA for cross-architecture and cross-domain black-box transferability.
\end{abstract}

\section{Introduction}
\label{sec:intro}

The emergence of Vision-Language-Action (VLA) models~\cite{openvla,Univla,cronusvla,instructvla,pi,objectVLA,dexvla} has bridged the gap between perception and control, enabling robots to leverage the reasoning capabilities of large models for complex physical interactions. 
However, as these models permeate safety-critical domains such as medical assistance and home services, their severe vulnerability to adversarial attacks significantly hinders their real-world deployment. Compared to traditional Vision-Language Models (VLMs), VLA models interact directly with the physical world, rendering their adversarial robustness significantly more critical~\cite{vlasafety1,vlasafety2}. While current research extensively investigates the adversarial vulnerabilities of VLMs~\cite{pang2023,Lingoloop,cmi-attack}, the security of VLA models has received comparatively limited attention~\cite{uada,patchper1}.

In the context of adversarial vulnerabilities, patch attacks have garnered significant attention due to their practical deployability~\cite{patchattack3,patchattack4,patchattack5}. Unlike traditional adversarial attacks that require imperceptible modifications across the entire image, patch attacks introduce localized, visible patterns within the camera's field of view, presenting a tangible security threat~\cite{patchattack1,pg-attack,patchattack2}.
Existing attacks on VLA models~\cite{uada} have primarily targeted the output action space, aiming to distort control signals by maximizing the action deviation. While effective in white-box settings, we identify a critical bottleneck in this paradigm: poor transferability. VLA models are characterized by heterogeneous action encodings and highly distinctive action space distributions that depend on the robot's physical morphology and training data. Consequently, adversarial perturbations that optimize for action divergence tend to overfit the specific "Action Head" of the source model~\cite{openvla,Univla,cronusvla}. This dependency renders the attack brittle, preventing generalization to unseen architectures with distinct decision-making.

\begin{wrapfigure}{r}{0.45\textwidth}
\centering
\vspace{-20pt}
\includegraphics[scale=0.2]{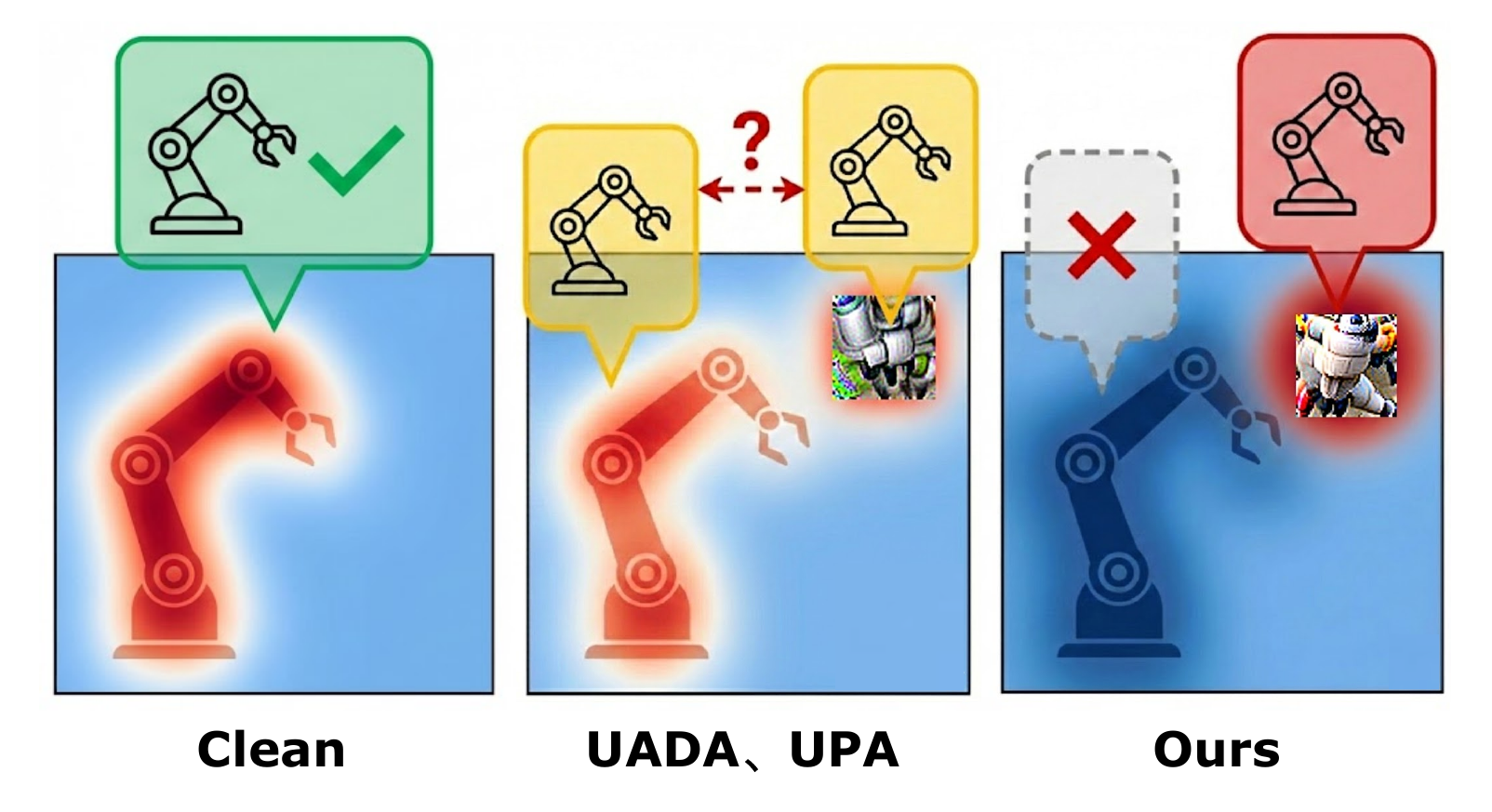} 
\vspace{-20pt}
\caption{Unlike prior attacks that leave the real arm unsuppressed, VLA-Hijack hijacks the proprioceptive loop by suppressing the original embodiment and injecting a surrogate identity.}
\vspace{-15pt}
\label{pic:intro}
\end{wrapfigure}

To break this transferability bottleneck, we propose a paradigm shift: targeting the universal proprioceptive logic of VLA models rather than their divergent action outputs. Intriguingly, recent adversarial studies~\cite{uada,patchper1,patchper2} have uncovered a provocative phenomenon: even when solely optimizing for action error, the generated adversarial patches unintentionally converge towards visual patterns resembling robotic arm joints (as visualized in the middle panel of Fig.~\ref{pic:intro}). While prior work attributes this to representation bias from restricted camera views~\cite{uada}, we argue this exposes a more fundamental vulnerability in VLA decision-making: the reliance on a \textbf{visual proprioceptive loop}. VLA manipulation is strictly contingent on scene perception: the model must \textbf{first localize its own physical state (the arm) from the current observation}, then locate the target object, and finally plan the relative motion~\cite{RT-2,openvla}. In this context, visual proprioception acts as the model's intrinsic perception and awareness of its own robotic morphology. Consequently, when a patch's features mimic the robotic arm, it effectively hijacks this internal perception, deceiving the model into misidentifying the patch as its true embodiment. Once this "self-localization" step is compromised, the model plans based on an erroneous physical origin (the patch location), inducing a systematic deviation in all subsequent trajectory calculations and leading to catastrophic failure. Since the capability to extract robotic arm semantics is universally acquired by VLAs during training, targeting this shared semantic layer fundamentally resolves the transferability challenges plaguing action-space attacks. This is empirically validated in Table~\ref{tab:2}, where a naive "Arm Image" consistently outperforms several sophisticated attacks. This confirms that robotic arm features represent a robust, universal focal point that VLA models inherently rely upon for decision-making.

Motivated by this insight, we propose VLA-Hijack, a unified adversarial framework designed to sever the relationship between the agent's embodiment and its control policy through a \textbf{Proprioceptive Identity Rewiring} mechanism. We optimize two complementary objectives: (1) \textit{Attention-Guided Proprioceptive Suppression}, which actively inhibits the feature responses of the original arm. Unlike prior strategies~\cite{uada} that suffer from severe semantic interference by failing to explicitly suppress original proprioceptive features, VLA-Hijack creates a "feature vacuum" that blinds the model to its true physical embodiment; and (2) \textit{Multimodal Proprioceptive Injection}, which fills this void by establishing the adversarial patch as a "phantom embodiment". By employing a dynamic schedule that alternates between semantic concept anchoring and visual prototype projection, we ensure the patch captures robust identity features, thereby addressing the indirect and suboptimal nature of existing optimization paths that rely solely on passive action error backpropagation~\cite{uada}. This synchronized manipulation effectively "hijacks" the model's self-localization capability, ensuring superior attack transferability across diverse VLA architectures.

In summary, our main contributions are as follows:
\begin{itemize}
    \item We uncover an explainable VLA decision logic: models must visually localize their robotic arm before planning motion. We further identify this mechanism as a novel insight for significantly enhancing attack transferability across heterogeneous VLA architectures. 
    \item We propose VLA-Hijack, a unified adversarial framework. By synergistically coupling attention-guided suppression with alternating multimodal injection, it synthesizes a phantom embodiment to hijack the model's decision-making.
    \item Evaluations across 12 victim models in the Real-to-Sim setting demonstrate that VLA-Hijack sets a new SOTA for transferability. Our method achieves a 63.68\% average Failure Rate, outperforming existing SOTA baselines by absolute margins of \textbf{42.66\%} to \textbf{45.41\%}.
    
\end{itemize}    

\section{Related Work}

\subsection{Vision-Language-Action Models}

Vision-Language-Action (VLA) models have emerged as powerful generalist policies by integrating robotic action generation into pretrained vision-language foundations~\cite{pi,instructvla,objectVLA,openvla,Univla,cronusvla}. A prominent line of work~\cite{RT-2,openvla,RoboFlamingo,spatialvla}, including RT-2~\cite{RT-2} and OpenVLA~\cite{openvla}, treats physical control as a language modeling task, discretizing continuous actions into tokens for autoregressive prediction. Conversely, to bypass the limitations of discrete formulations and the heavy reliance on labeled interactive data, recent advancements explore continuous action heads and latent representations~\cite{Univla,cronusvla,tinyvla,vla_undiscrete1,instructvla}. For instance, UniVLA~\cite{Univla} learns a unified latent action representation from internet-scale action-free videos, while CronusVLA~\cite{cronusvla} adapts single-frame discrete policies to a multi-frame continuous paradigm. Given the significant heterogeneity in their action space designs—spanning discrete tokenization, unified latent representations, and continuous prediction—we select OpenVLA~\cite{openvla}, UniVLA~\cite{Univla}, and CronusVLA~\cite{cronusvla} as our representative targets to evaluate the cross-architecture transferability of our proposed attack.

\vspace{-6pt}

\subsection{Attacks on VLA Models}
As VLA models are increasingly deployed in embodied applications, their adversarial vulnerabilities have garnered significant attention. A substantial body of work investigates training-time threats, demonstrating how VLA systems can be compromised through data poisoning and backdoor attacks~\cite{backdoor1,backdoor2,backdoor3,backdoor4,backdoor5,backdoor6,backdoor7}. While these methods embed malicious behaviors by altering the training or fine-tuning pipelines, another line of research focuses on inference-time evasion. Within this scope, initial adversarial attacks typically employ full-image perturbations, applying imperceptible noise across the entire visual observation to mislead the model's execution~\cite{fullper1,fullper2}. To present a more realistic threat vector without modifying every pixel, patch attacks introduce localized, visible patterns within the camera's field of view~\cite{patchper1,uada,patchper2}.
However, existing patch attacks on VLA models often suffer from poor cross-architecture transferability. To address this limitation, we propose a novel adversarial framework that significantly improves transferability by explicitly hijacking the model's decision-making through the suppression of its visual proprioception.

\vspace{-6pt}

\section{Methodology}


\subsection{Preliminaries \& Threat Model}
In this section, we formally define the VLA model and its inference mechanism, followed by the formulation of our adversarial attack objective.
\vspace{-5pt}
\paragraph{VLA Formulation.}
We define a VLA model as a parameterized function $f_\theta$, designed to generate physical control signals based on visual observations and linguistic instructions. Specifically, given an RGB observation image $I \in \mathbb{R}^{H \times W \times 3}$ and a natural language instruction $T$ (e.g., \textit{``Pick up the blue cube''}), the inference process of a VLA model is formulated as:
\begin{equation}
    A = f_\theta(I, T),
\end{equation}
where $A$ denotes the generated action trajectories.
\vspace{-5pt}
\paragraph{Attack Objective.}
This work focuses on \textit{universal untargeted attacks} against VLA models—meaning a single perturbation can generalize across various tasks. We aim to generate a single adversarial patch $P$ that, applied across diverse observations, causes task failure by inducing erroneous actions without altering the instruction $T$.
Formally, we define the adversarial observation image $I_{adv}$ as:
\begin{equation}
    I_{adv} = I \odot (1-M) + P \odot M ,
    \label{eq:adv_img}
\end{equation}
where $M \in \{0, 1\}^{H \times W}$ represents the binary mask of the patch (with 1 indicating the patch region and 0 otherwise), and $\odot$ denotes element-wise multiplication. 

To achieve the attack goal, we formulate the patch generation as a constrained optimization problem. We seek to find the optimal patch $P$ that minimizes a specific adversarial objective function $\mathcal{L}_{total}$ over the data distribution $\mathcal{D}$:
\begin{equation}
    \min_{P} \mathbb{E}_{(I, T) \sim \mathcal{D}} \left[ \mathcal{L}_{total}(f_\theta(I_{adv}, T)) \right] \quad \text{s.t.} \quad P \in [0, 1]^{H \times W} , 
    \label{eq:optimization_objective}
\end{equation}
where $\mathbb{E}_{(I, T) \sim \mathcal{D}}$ denotes the expectation over the data distribution $\mathcal{D}$, ensuring the generated patch $P$ achieves universal effectiveness across diverse visual observations and instructions. In our proposed framework, this objective is instantiated as a unified loss function $\mathcal{L}_{total}$, as detailed in Sec.~\ref{sec:optimization}.

\paragraph{Threat Model.}
We adopt a \textbf{transfer-based attack} setting, where the attacker is assumed to have white-box access to a single \textit{surrogate model} $f_{\theta}$ to compute gradients and optimize the patch $P$. During the inference phase, the optimized patch is evaluated against other unseen \textit{victim models} in a black-box manner. In this setting, the attacker has no knowledge of the victim models' internal parameters or architectures, relying solely on the cross-model transferability of the adversarial features captured by $P$.

\begin{figure}[t]
    \centering
    \includegraphics[width=0.65\linewidth]{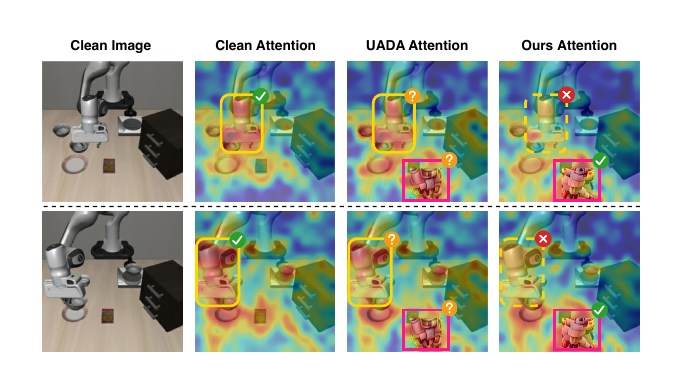}
    \vspace{-10pt}
    \caption{\textbf{Visualization of Proprioceptive Conflict.} Unlike UADA (Col 3), which splits attention between the real arm and the patch, our method (Col 4) completely suppresses the real arm and redirects focus exclusively to the adversarial patch.}
    \vspace{-18pt}
    \label{fig:motivation}
\end{figure}

\vspace{-8pt}

\subsection{Overview of VLA-Hijack}
\subsubsection{Motivation: The Proprioceptive Conflict.} Our framework is driven by a critical phenomenon observed in the attention mechanisms of VLA models. As illustrated in Fig. \ref{fig:motivation}, VLA manipulation is strictly contingent on a \textbf{"Visual Proprioception Loop"}: the model must first localize its own physical state (the robotic arm) from the current observation, then locate the target object, and finally plan the robotic arm's relative motion. In the Clean State (Fig. \ref{fig:motivation}, Col 2), the attention map clearly highlights both the \textit{target object} and the \textit{robotic arm}, confirming that self-localization is a prerequisite for control.

\begin{figure}[t]
    \hspace*{-0.5cm}\centering
    \includegraphics[width=1.1\linewidth]{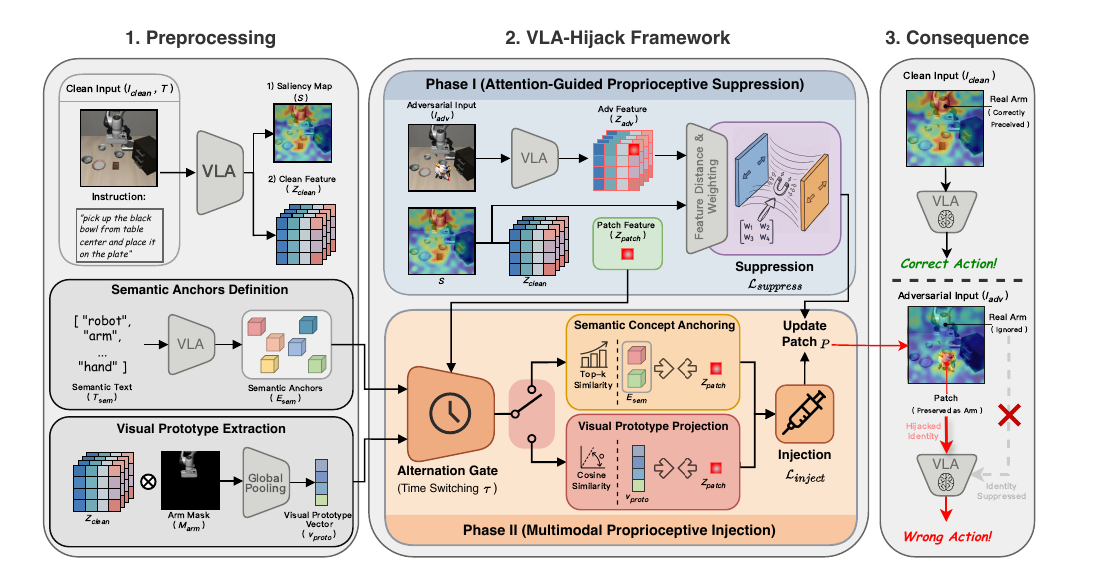}
    \vspace{-20pt}
    
    \caption{\textbf{Overview of the VLA-Hijack Framework.}We employ a two-phase optimization loop to generate the adversarial patch: first suppressing the features of the real arm, and then injecting multimodal robotic arm identities into the patch.}
    \vspace{-14pt}
    \label{fig:pipeline}
\end{figure}


Crucially, prior attacks (e.g., UADA)~\cite{uada} inadvertently reveal this dependency. As shown in Fig.~\ref{fig:motivation}, even when relying on indirect objectives like action-error backpropagation rather than direct perceptual guidance, adversarial patches spontaneously evolve into patterns resembling robotic components. While this confirms that VLA decisions are fundamentally anchored to visual self-localization, such indirect optimization is inherently suboptimal. By failing to suppress the \textit{original embodiment}, these methods result in a severe \textbf{"Semantic Conflict"} where the model's attention is split between the salient real arm and the patch. To achieve robust and transferable control, we must therefore address this bottleneck directly: we need to hijack the proprioceptive loop by removing the real embodiment and establishing the patch as the sole perceived embodiment.

\vspace{-10pt}

\subsubsection{The VLA-Hijack Framework.}
To address this bottleneck, we propose \textbf{VLA-Hijack}, a unified framework designed to execute the aforementioned identity replacement. As illustrated in Fig. \ref{fig:pipeline}, the framework operates through a synergistic \textbf{``Proprioceptive Identity Rewiring''} mechanism, explicitly targeting the visual proprioception loop via two concurrent objectives:
\vspace{-4pt}

\begin{itemize} 
    \item \textit{Attention-Guided Proprioceptive Suppression.} We penalize the feature response of the real robotic arm. This creates the "perceptual vacuum" (visualized as the blue region in Fig. \ref{fig:motivation}, Col 4), blinding the model to its true physical presence. 
    \item \textit{Multimodal Proprioceptive Injection.} Synchronized with the suppression, we inject the visual and semantic prototypes of a target embodiment into the patch. This forces the model to "latch onto" the patch (visualized as the red region in Fig. \ref{fig:motivation}, Col 4) as its sole embodiment. 
\end{itemize} 


\vspace{-10pt}

\subsection{Phase I: Attention-Guided Proprioceptive Suppression}\label{sec:phase1}

To sever the relationship between the model and its  embodiment, this phase exploits the VLA model's internal attention mechanisms to identify and suppress proprioceptive features. 

\vspace{-4pt}

\paragraph{Saliency-Guided Weighting.}
We directly extract the multi-head self-attention maps from the final layer of the visual encoder. By aggregating the attention weights across all heads, we obtain a raw spatial saliency map $S \in \mathbb{R}^{h \times w}$, where $h$ and $w$ denote the spatial dimensions of the feature map.
Given that the high-response regions in $S$ capture the model's focus on proprioceptive features (i.e., the robotic arm), we transform $S$ into a penalty weight map $W$ via normalization:
\begin{equation}
    W = \mathbf{1} + \alpha \cdot \frac{S - S_{min}}{S_{max} - S_{min} + \epsilon}~~,
    \label{eq:weight_map}
\end{equation}
where $S_{min}$ and $S_{max}$ denote the minimum and maximum values of the saliency map, $\epsilon$ ensures numerical stability, and $\alpha$ is a scaling factor that amplifies the penalty intensity on these high-attention regions.

\vspace{-4pt}

\paragraph{Weighted Feature Suppression.}Leveraging $W$, we introduce a weighted loss to eliminate the original semantic information. Let $Z_{clean}$ and $Z_{adv}$ denote the feature maps of the clean and adversarial images, respectively. We minimize their weighted cosine similarity:
\begin{equation}
\mathcal{L}_{suppress} = \frac{\sum (W \odot \text{CosSim}(Z_{adv}, Z_{clean}))}{\sum W}~~,
\label{eq:erase_loss}
\end{equation}
Crucially, this explicit weighting prevents the optimizer from trivially minimizing the loss by perturbing unstable background tokens, forcing it to prioritize dismantling the highly robust proprioceptive features.
By minimizing $\mathcal{L}_{suppress}$, we force the adversarial features to deviate orthogonally from the original features, specifically in regions where the model originally attended, effectively creating a "perceptual vacuum" around the robotic arm.

\vspace{-6pt}

\subsection{Phase II: Multimodal Proprioceptive Injection}
\label{sec:phase2}

Following the suppression of original features, this phase aims to fill the "feature vacuum" by synthesizing a surrogate embodiment. To this end, we formulate two complementary injection objectives: \textbf{Semantic Concept Anchoring} and \textbf{Visual Prototype Projection}.
\vspace{-4pt}

\paragraph{Semantic Concept Anchoring.}
To activate the language alignment modules within the VLA model, we anchor the patch features to a predefined semantic space. 
We define a set of text prompts $\mathcal{T}_{sem}$ describing the embodiment (e.g., \textit{\{"robot", "arm", "hand", "claw", "end-effector"\}}) and feed them into the text encoder to obtain normalized embeddings $E_{sem} \in \mathbb{R}^{K \times D}$, serving as \textbf{Semantic Anchors}.
To enforce strong semantic correlation while addressing the inherent semantic polysemy of robotic features (e.g., distinct textures resembling a "claw" versus an "arm"), we reject rigid one-to-one mapping in favor of a flexible Top-$k$ Aggregation Strategy. We compute the cosine similarity between the adversarial patch features and all anchors, maximizing the average similarity to the top-$k$ most relevant ones for each token:
\begin{equation}
    \mathcal{L}_{sem} = 1 - \frac{1}{|\Omega|} \sum_{j \in \Omega} \left( \frac{1}{k} \sum_{e \in \text{TopK}(Z_{adv}^{(j)}, E_{sem})} \frac{Z_{adv}^{(j)} \cdot e}{\|Z_{adv}^{(j)}\|_2 \|e\|_2} \right),
    \label{eq:loss_text}
\end{equation}
where $\Omega$ denotes the set of spatial token indices covering the adversarial patch region, and $|\Omega|$ represents the total number of patch tokens. $Z_{adv}^{(j)}$ is the $j$-th token's feature vector, and $\text{TopK}(\cdot)$ selects the $k$ anchors with the highest similarity scores. By aggregating these top candidates, this objective forces the patch features to converge towards a comprehensive semantic centroid of the "robotic arm" concept, ensuring robust alignment across diverse visual attributes.

\vspace{-4pt}

\paragraph{Visual Prototype Projection.}
While semantic anchoring ensures the patch is recognized as a generic "robotic arm" in the linguistic space, it fails to capture the fine-grained visual characteristics (e.g., specific textures, lighting, and geometric details) of the actual embodiment. To bridge this modality gap and achieve a holistic identity injection, we introduce \textbf{Visual Prototype Projection} to directly clone the visual features of the physical arm.

Prior to feature extraction, we explicitly isolate the physical embodiment region. We utilize the segmentation model to obtain the binary mask $M_{arm}$ corresponding to the robotic arm from the clean observation $I_{clean}$.

Based on this mask, we extract a robust \textbf{Visual Prototype Vector} $v_{proto} \in \mathbb{R}^{D}$ by performing global average pooling over the masked region of the clean features $Z_{clean}$, followed by $\ell_2$ normalization:
\begin{equation}
    v_{proto} = \frac{\sum_{i} M_{arm}^{(i)} \cdot Z_{clean}^{(i)}}{\left\| \sum_{i} M_{arm}^{(i)} \cdot Z_{clean}^{(i)} \right\|_2 + \epsilon}~~.
    \label{eq:v_proto}
\end{equation}
Subsequently, we define the \textbf{Visual Injection Loss} $\mathcal{L}_{vis}$ as the average cosine distance between the adversarial features in the patch and the prototype:
\begin{equation}
    \mathcal{L}_{vis} = 1 - \frac{1}{|\Omega|} \sum_{j \in \Omega} \frac{Z_{adv}^{(j)} \cdot v_{proto}}{\|Z_{adv}^{(j)}\|_2 \|v_{proto}\|_2}~~.
    \label{eq:loss_vis}
\end{equation}
Minimizing $\mathcal{L}_{vis}$ ensures that the patch effectively ``clones'' the texture and structural information of the real robot arm in the feature space.

\vspace{-8pt}

\subsection{Optimization}
\label{sec:optimization}

We formulate the attack generation as a joint optimization problem. Our ultimate goal is to find an optimal patch $P$ that simultaneously maintains a "perceptual vacuum" on the original embodiment while firmly establishing the surrogate identity. The total objective function is defined as:
\begin{equation}
    \min_{P} \mathcal{L}_{total} = \mathcal{L}_{suppress} + \lambda \cdot \mathcal{L}_{inject}^{(t)}~~,
\end{equation}
where $\lambda$ is a hyperparameter balancing suppression and injection strength.
\vspace{-4pt}

\paragraph{Alternating Injection Schedule.}
While the suppression term remains constant, strictly optimizing both injection modalities simultaneously can lead to gradient conflicts. To address this, we define the injection term $\mathcal{L}_{inject}^{(t)}$ as a dynamic component that alternates between semantic and visual modalities based on the iteration step $t$ and a period $\tau$:
\begin{equation}
    \mathcal{L}_{inject}^{(t)} = 
    \begin{cases} 
    \mathcal{L}_{sem}(Z_{adv}, E_{sem}), & \text{if } \lfloor \frac{t}{\tau} \rfloor \text{ is even} \\
    \mathcal{L}_{vis}(Z_{adv}, v_{proto}), & \text{if } \lfloor \frac{t}{\tau} \rfloor \text{ is odd} 
    \end{cases}~~.
\end{equation}
This alternating strategy mitigates optimization conflicts, allowing the patch to iteratively align with the target identity across modalities while continuously neutralizing the real arm via $\mathcal{L}_{suppress}$.
Full procedures are in Supplement.

\vspace{-4pt}

\section{Experiments}

\subsection{Experiment Setup}

\subsubsection{Datasets and Baseline.}
To comprehensively evaluate our attack framework, we conduct experiments on LIBERO~\cite{libero} and BridgeData V2~\cite{bridgedata}. LIBERO~\cite{libero} serves as our primary simulation benchmark, assessing high-level reasoning through four task suites: Spatial, Object, Goal, and Long. These suites challenge models with complex, multi-step sequential decision-making. Complementing this simulation environment, BridgeData V2~\cite{bridgedata} provides a large-scale real-world corpus comprising 60,096 trajectories across 24 diverse environments. Covering 13 fundamental manipulation skills, it allows us to rigorously test the VLA robustness in unstructured physical scenes. We benchmark against 8 adversarial objectives~\cite{uada} (e.g., UADA, UPA, and TMA with varying Degrees-of-Freedom), as well as Random Noise and Arm Image to evaluate naive visual priors.

\vspace{-10pt}

\subsubsection{Victim VLAs.}
We evaluate adversarial robustness across three representative VLA architectures: OpenVLA~\cite{openvla}, UniVLA~\cite{Univla}, and CronusVLA~\cite{cronusvla}. In the LIBERO simulation domain~\cite{libero}, we explicitly utilize four distinct variants for each architecture, independently trained on the specific task suites (i.e., Spatial, Object, Goal, and Long), resulting in a total of 12 distinct victim models. For white-box optimization, we employ OpenVLA~\cite{openvla} as the primary surrogate architecture. Specifically, we select the OpenVLA-Spatial variant, as it exhibits the lowest failure rate on clean samples, thereby ensuring that attack success is attributed strictly to adversarial efficacy rather than the intrinsic fragility of the surrogate. To further assess cross-architecture transferability, we extend our surrogate set to include all four task-specific variants of UniVLA~\cite{Univla}. In the physical domain, regarding the BridgeData V2~\cite{bridgedata}, we utilize the OpenVLA-7B model to evaluate performance in unstructured environments.

\vspace{-10pt}

\subsubsection{Attacks Settings and Metrics.}
For the attack configuration, we initialize the adversarial patch to cover $5\%$ of the input image area, aligning with baseline~\cite{uada}. To isolate the robotic arm, we extract the binary mask $M_{arm}$ using the SAM 3~\cite{sam3}. The optimization process is conducted with a batch size of $4$, spanning a total of Attack Steps $N=500$. We empirically set the saliency scaling factor $\alpha=2$ to enforce focused suppression on embodiment features. Within each step, we perform Inner-Loop Steps $K=50$ to ensure sufficient feature convergence. For evaluation, we follow the LIBERO benchmark~\cite{libero} and adopt the Failure Rate (FR) as our primary metric, defined as $1 - \text{Success Rate (SR)}$. 

\vspace{-10pt}

\subsubsection{Evaluation Details.}
To ensure the statistical reliability of our results, we adopt a comprehensive evaluation protocol on the LIBERO benchmark~\cite{libero}. Each task suite comprises 10 distinct manipulation tasks. For our main experiments, we conduct 50 independent rollout trials for every task to account for environmental stochasticity, yielding a robust total of 500 evaluation episodes per suite. Conversely, for all \textbf{ablation studies}, we streamline the protocol to 10 independent rollouts per task to efficiently manage computational overhead.

\begin{table}[t]
\centering
\fontsize{5.5pt}{8.2pt}\selectfont
\caption{\textbf{Attack effectiveness within the Simulation Domain.} We report the Failure Rate (FR, \%; $\uparrow$) in the LIBERO simulator. Patches are generated using the \textit{OpenVLA-Spatial} surrogate model and evaluated on 12 victim models. Shaded entries denote the white-box performance, while \textit{Transfer Avg.} reports the mean FR across the remaining 11 black-box variants.}
\label{tab:1}
\vspace{-6pt}
\begin{tabular}{@{}c|cccc|cccc|cccc|c@{}}
\toprule
    \multirow{2}{*}{Method} & \multicolumn{4}{c|}{OpenVLA}       & \multicolumn{4}{c|}{UniVLA}      & \multicolumn{4}{c|}{CronusVLA}   & \multicolumn{1}{l}{\multirow{2}{*}{\makecell{Transfer \\ Avg.}}} \\ \cmidrule(lr){2-13}
                        & Spatial* & Object & Goal   & Long  & Spatial & Object & Goal  & Long  & Spatial & Object & Goal  & Long  & \multicolumn{1}{l}{}                               \\ \midrule
Benign                  & 16.00    & 16.80  & 24.20  & 45.20 & 2.40    & 5.60   & 5.20  & 7.00  & 7.40    & 7.60   & 4.20  & 13.80 & -                                              \\
Random Noise            & 26.80    & 27.40  & 24.60  & 48.60 & 3.80    & 9.60   & 6.60  & 7.40  & 11.80   & 7.60   & 4.60  & 14.20 & 15.11                                              \\
Arm image               & 66.20    & 44.40  & 28.80  & 60.80 & 4.80    & 11.20  & 9.40  & 15.80 & 8.20    & 11.40  & 5.20  & 15.20 & 19.56                                              \\ \midrule
UADA$_1$                & \cellcolor{gray!38}38.80    & 37.00  & 26.40  & 53.80 & 6.80    & 9.40   & 8.40  & 13.20 & 15.60   & 9.00   & 5.80  & 15.40 & 18.25                                              \\
UADA$_{1-3}$            & \cellcolor{gray!38}\textbf{100.00}   & 92.20  & 99.80  & 96.60 & 9.60    & 24.80  & 15.00 & 24.00 & 17.80   & 8.80   & 6.20  & 18.00 & 37.53                                              \\ \midrule
UPA$_1$                 & \cellcolor{gray!38}62.20    & 47.20  & 32.20  & 59.60 & 6.80    & 14.00  & 7.60  & 21.40 & 12.80   & 10.20  & 5.40  & 16.40 & 21.24                                              \\
UPA$_{1-3}$             & \cellcolor{gray!38}55.40    & 44.20  & 30.60  & 65.20 & 6.00    & 11.80  & 9.80  & 17.80 & 14.40   & 9.40   & 4.80  & 14.60 & 20.78                                              \\ \midrule
TMA$_1$                 & \cellcolor{gray!38}99.40    & 79.20  & 44.00  & 80.20 & 10.80   & 14.60  & 8.60  & 29.00 & 14.40   & 8.80   & 5.60  & 17.20 & 28.40                                              \\
TMA$_{1-3}$             & \cellcolor{gray!38}99.00    & 82.20  & 52.00  & 84.20 & 32.60   & 19.60  & 8.00  & 30.60 & 15.40   & 8.40   & 5.80  & 15.80 & 32.24                                              \\
TMA$_7$                 & \cellcolor{gray!38}98.00    & 70.00  & 40.60  & 82.40 & 6.80    & 10.20  & 9.60  & 27.60 & 17.40   & 6.60   & 4.60  & 16.80 & 26.60                                              \\
TMA$_{1-7}$             & \cellcolor{gray!38}96.20    & 92.80  & 46.60  & 83.80 & 14.60   & 11.40  & 9.00  & 14.60 & 13.40   & 9.60   & 5.60  & 14.60 & 28.73                                              \\ \midrule
Ours                    & \cellcolor{gray!38}\textbf{100.00}   & \textbf{100.00} & \textbf{100.00} & \textbf{99.60} & \textbf{88.20}   & \textbf{100.00} & \textbf{25.20} & \textbf{80.80} & \textbf{28.20}   & \textbf{12.40}  & \textbf{25.80} & \textbf{17.80} & \textbf{61.64}                                              \\ \bottomrule
\end{tabular}
\vspace{-10pt}
\end{table}

\vspace{-6pt}

\subsection{Main Results}
\subsubsection{Attack Effectiveness within the Simulation Domain.}
As reported in Table~\ref{tab:1}, our proposed VLA-Hijack achieves state-of-the-art performance in the white-box setting, attaining a perfect 100.00\% Failure Rate (FR) on the surrogate model (OpenVLA-Spatial). Simultaneously, our method demonstrates superior transferability across unseen architectures, significantly outperforming existing attacks.
Since prior methods largely neglect the model's fundamental dependency on the visual proprioceptive loop, they tend to overfit the surrogate model, resulting in severe performance degradation on diverse victim models (e.g., the strongest baseline UADA$_{1-3}$ averages only 37.53\% in transfer FR). In contrast, by successfully capturing and exploiting the universal embodiment features shared across heterogeneous VLA models, VLA-Hijack bridges this gap, achieving a remarkable Transfer Avg. of 61.64\%.

\vspace{-10pt}

\subsubsection{Robustness to Visual Domain Shifts (Real-to-Sim).}
To further rigorously evaluate robustness under severe distribution shifts, we report the transfer performance in Table~\ref{tab:2}, where adversarial patches are trained on real-world BridgeData V2 and tested on LIBERO simulation. This setting poses a significant challenge due to the substantial "visual domain gap" (e.g., lighting, textures) between physical and simulation environments. Consequently, prior optimization-based attacks suffer catastrophic failure, yielding an average FR of only 18\%-21\%, which is surprisingly even lower than the naive "Arm image" baseline (23.45\%). This indicates that their generated features are highly sensitive to domain-specific statistics. In stark contrast, VLA-Hijack effectively bridges this domain gap, achieving a commanding Transfer Avg. of 63.68\%. By anchoring the attack on the semantic concept of the "robotic arm"—a visual feature that remains invariant across real and simulated worlds—our method realizes a 3$\times$ performance improvement over state-of-the-art baselines, demonstrating that the "phantom embodiment" we synthesize possesses true cross-domain universality.

\begin{table}[t]
\centering
\fontsize{5.5pt}{8.1pt}\selectfont
\caption{\textbf{Cross-Domain Evaluation (Real $\to$ Sim).} We report the FR (\%) in the LIBERO simulator. Patches are generated from the real-world BridgeData V2 dataset using \textit{OpenVLA-7B} as the surrogate model, and transferred to 12 simulation models. \textit{Transfer Avg.} reports the mean FR over all 12 victims.}
\vspace{-8pt}
\label{tab:2}
\begin{tabular}{@{}c|cccc|cccc|cccc|c@{}}
\toprule
\multirow{2}{*}{Method} & \multicolumn{4}{c|}{OpenVLA}       & \multicolumn{4}{c|}{UniVLA}      & \multicolumn{4}{c|}{CronusVLA}   & \multicolumn{1}{l}{\multirow{2}{*}{\makecell{Transfer \\ Avg.}}} \\ \cmidrule(lr){2-13}
                        & Spatial & Object & Goal  & Long  & Spatial & Object & Goal  & Long  & Spatial & Object & Goal  & Long  &                               \\ \midrule
Benign                  & 16.00   & 16.80  & 24.20 & 45.20 & 2.40    & 5.60   & 5.20  & 7.00  & 7.40    & 7.60   & 4.20  & 13.80 & -                             \\
random Noise            & 26.80   & 27.40  & 24.60 & 48.60 & 3.80    & 9.60   & 6.60  & 7.40  & 11.80   & 7.60   & 4.60  & 14.20 & 16.08                         \\
Arm image               & 66.20   & 44.40  & 28.80 & 60.80 & 4.80    & 11.20  & 9.40  & 15.80 & 8.20    & 11.40  & 5.20  & 15.20 & 23.45                         \\ \midrule
UADA$_1$              & 32.00   & 32.40  & 26.80 & 50.80 & 5.20    & 10.40  & 7.80  & 9.60  & 14.00   & 9.20   & 5.60  & 15.40 & 18.27                         \\
UADA$_{1-3}$            & 29.40   & 36.20  & 28.40 & 49.40 & 4.60    & 11.00  & 10.40 & 11.40 & 15.00   & 8.40   & 6.20  & 13.80 & 18.68                         \\ \midrule
UPA$_{1}$               & 29.00   & 34.60  & 25.80 & 48.60 & 5.40    & 11.80  & 9.80  & 9.20  & 14.20   & 10.20  & 6.40  & 15.40 & 18.37                         \\
UPA$_{1-3}$             & 31.20   & 32.40  & 28.00 & 50.60 & 5.40    & 12.60  & 8.60  & 9.00  & 15.00   & 10.00  & 6.00  & 15.40 & 18.68                         \\ \midrule
TMA$_{1}$               & 30.40   & 38.40  & 31.40 & 51.20 & 4.60    & 10.40  & 8.80  & 13.40 & 13.80   & 8.60   & 5.00  & 14.20 & 19.18                         \\
TMA$_{1-3}$             & 30.80   & 39.00  & 30.00 & 54.00 & 6.60    & 8.60   & 9.80  & 10.40 & 14.00   & 9.80   & 4.80  & 14.60 & 19.37                         \\
TMA$_{7}$               & 37.80   & 36.80  & 27.80 & 52.00 & 5.20    & 6.00   & 8.20  & 8.20  & 15.20   & 10.60  & 5.40  & 15.20 & 19.03                         \\
TMA$_{1-7}$             & 36.40   & 41.80  & 31.80 & 51.80 & 8.00    & 9.60   & 9.40  & 14.60 & 16.00   & 11.00  & 5.80  & 16.00 & 21.02                         \\ \midrule
Ours                    & \textbf{100.00} & \textbf{99.80} & \textbf{99.80} & \textbf{99.40} & \textbf{93.40} & \textbf{100.00} & \textbf{17.80} & \textbf{82.80} & \textbf{26.20} & \textbf{13.80} & \textbf{11.40} & \textbf{19.80} & \textbf{63.68}                \\ \bottomrule
\end{tabular}
\end{table}

\begin{table}[t]
\centering
\fontsize{5.5pt}{8.2pt}\selectfont

\caption{\textbf{Robustness to Surrogate Selection.} We report FR (\%) in the LIBERO simulator. Patches are generated using each \textit{UniVLA} variant as the surrogate and evaluated across all 12 victim models. Shaded entries denote white-box performance, while \textit{Transfer Avg.} reports the mean FR over the remaining 11 transfer victims.}
\label{tab:3}
\vspace{-6pt}
\begin{tabular}{@{}c|c|cccc|cccc|cccc|c@{}}
\toprule
\multirow{2}{*}{Method} & Victim$\rightarrow$ & \multicolumn{4}{c|}{UniVLA}    & \multicolumn{4}{c|}{OpenVLA}  & \multicolumn{4}{c|}{CronusVLA} & \multirow{2}{*}{\makecell{Transfer \\ Avg.}} \\ \cmidrule(lr){2-14}
                        & Surrogate$\downarrow$  & Spa.  & Obj.   & Goal  & Long  & Spa.  & Obj.  & Goal  & Long  & Spa.   & Obj.  & Goal  & Long  &                    \\ \midrule
Benign                  & \multirow{2}{*}{-}  & 2.40  & 5.60   & 5.20  & 7.00  & 16.00 & 16.80 & 24.20 & 45.20 & 7.40   & 7.60  & 4.20  & 13.80 & -                  \\
Rand. Noise             &                     & 3.80  & 9.60   & 6.60  & 7.40  & 26.80 & 27.40 & 24.60 & 48.60 & 11.80  & 7.60  & 4.60  & 14.20 & -                  \\ \midrule
\multirow{4}{*}{Ours}   & Uni-Spa.         & \cellcolor{gray!38}96.00 & \textbf{100.00} & 29.80 & 91.40 & 87.20 & \textbf{99.00} & 66.60 & 75.80 & \textbf{25.20}  & 12.60 & 20.40 & 19.20 & 57.02              \\
                        & Uni-Obj.         & \textbf{97.60} & \cellcolor{gray!38}\textbf{100.00} & \textbf{78.00} & 71.20 & \textbf{99.80} & 98.80 & \textbf{90.20} & 94.60 & 21.80  & 13.20 & \textbf{24.80} & 17.80 & \textbf{64.35}              \\
                        & Uni-Goal         & 52.80 & \textbf{100.00} & \cellcolor{gray!38}63.80 & 61.40 & 75.80 & 92.60 & 62.80 & 80.60 & 15.80  & \textbf{20.80} & 8.20  & 12.20 & 53.00              \\
                        & Uni-Long         & 82.60 & 97.20  & 53.60 & \cellcolor{gray!38}\textbf{96.00} & 98.80 & 91.00 & 84.00 & \textbf{99.00} & 21.60  & 13.80 & 8.40  & \textbf{20.80} & 60.98              \\ \bottomrule
\end{tabular}
\vspace{-6pt}

\end{table}

\begin{figure}[t]
    \centering
    
    \begin{subfigure}[b]{0.49\textwidth}
        \centering
        \includegraphics[width=\linewidth]{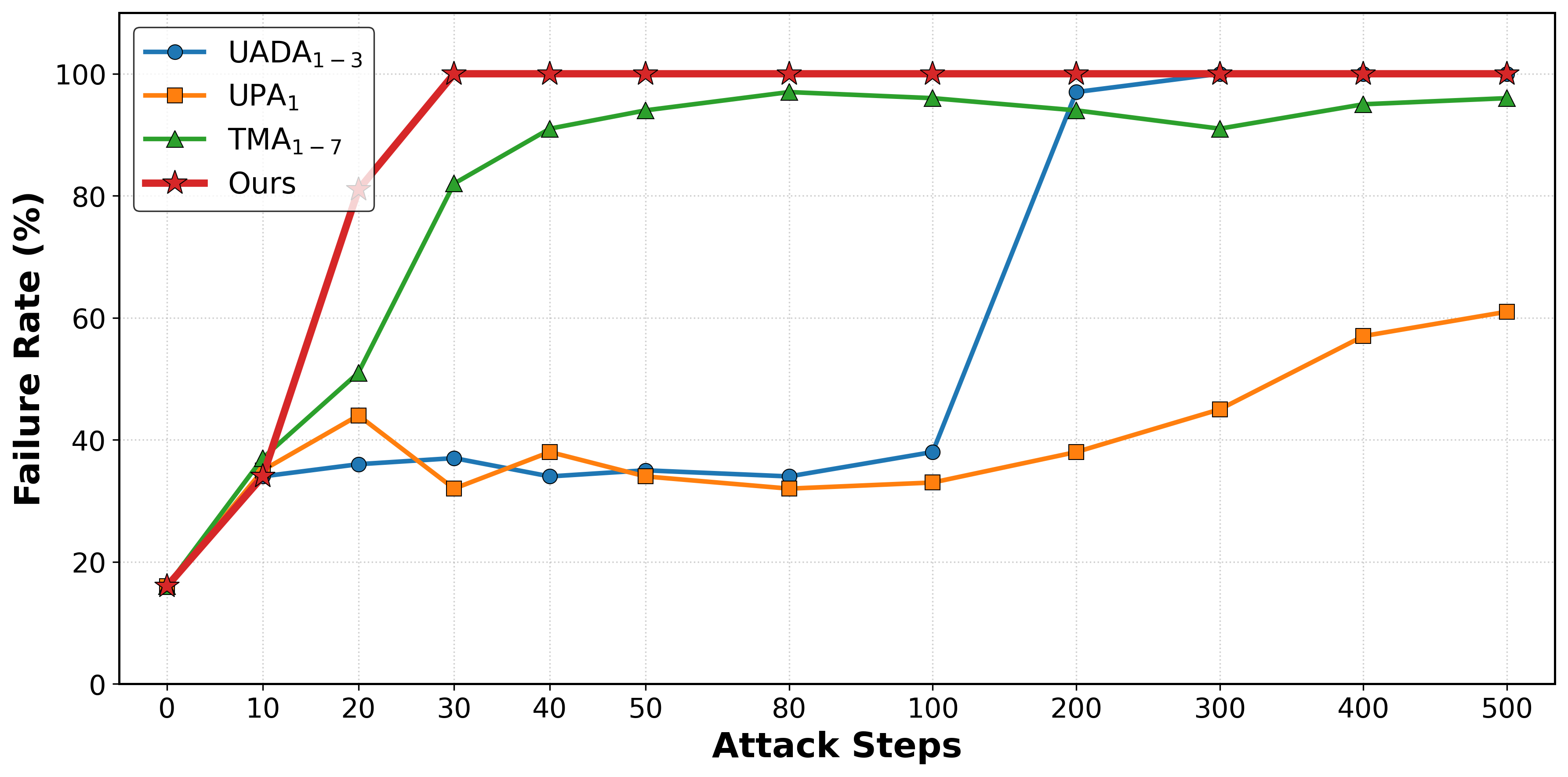} 
        \caption{\centering \textbf{Victim:} OpenVLA-Spatial \\ (Same Model, Same Task)}
        \label{fig:sub1}
    \end{subfigure}
    \hfill 
    \begin{subfigure}[b]{0.49\textwidth}
        \centering
        \includegraphics[width=\linewidth]{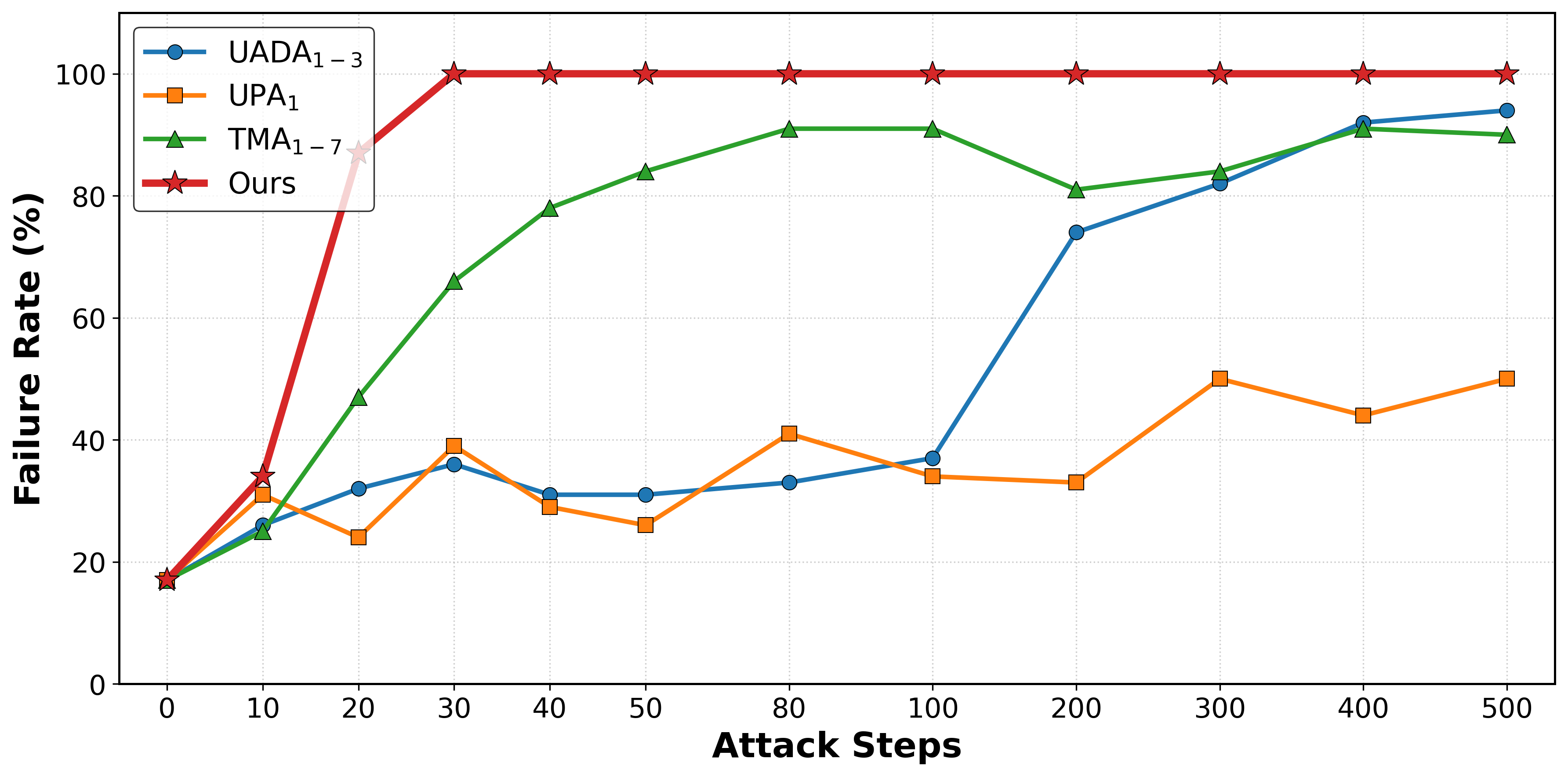}
        \caption{\centering \textbf{Victim:} OpenVLA-Object \\ (Same Model, Diff. Task)}
        \label{fig:sub2}
    \end{subfigure}
    
    
    \begin{subfigure}[b]{0.49\textwidth}
        \centering
        \includegraphics[width=\linewidth]{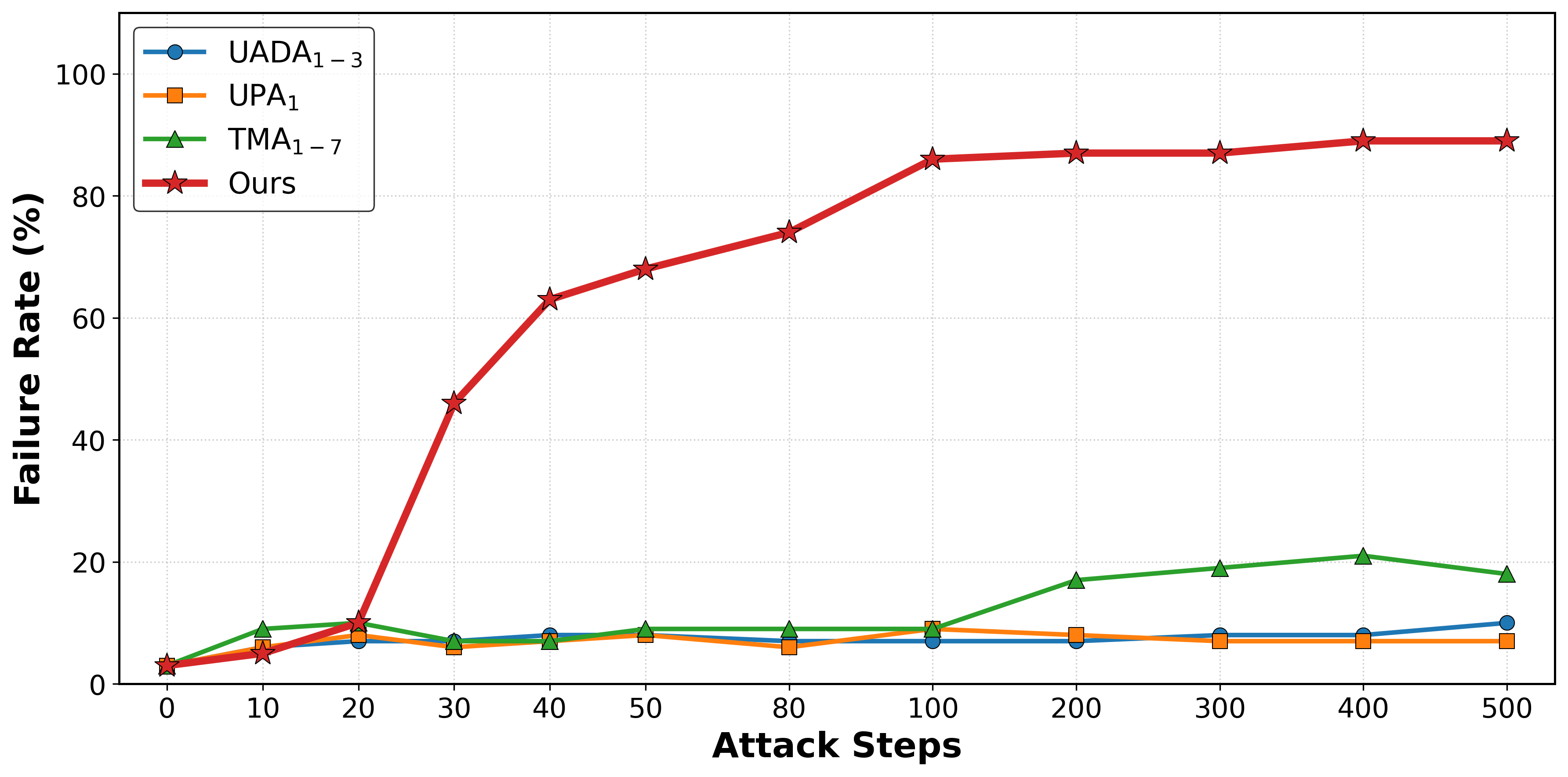}
        \caption{\centering \textbf{Victim:} UniVLA-Spatial \\ (Diff. Model, Same Task)}
        \label{fig:sub3}
    \end{subfigure}
    \hfill
    \begin{subfigure}[b]{0.49\textwidth}
        \centering
        \includegraphics[width=\linewidth]{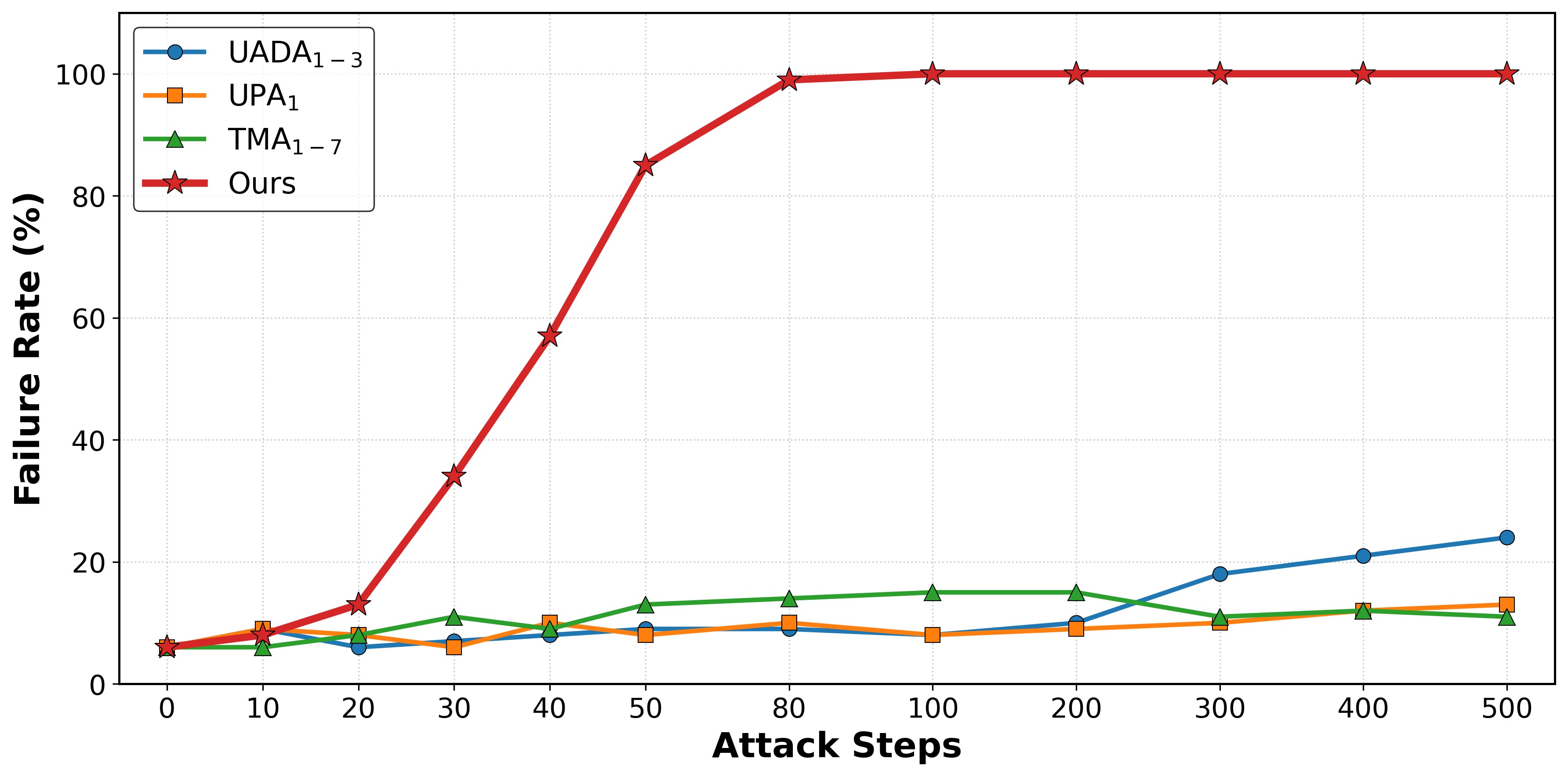}
        \caption{\centering \textbf{Victim:} UniVLA-Object \\ (Diff. Model, Diff. Task)}
        \label{fig:sub4}
    \end{subfigure}
    \caption{\textbf{Failure Rate (FR) vs. Optimization Steps.} We plot FR against attack steps using patches from the \textit{OpenVLA-Spatial} surrogate. Compared to baselines, \textbf{VLA-Hijack} (red curve) demonstrates faster convergence in the white-box setting (a) and continuous performance gains in transfer attacks (b-d).}
    \vspace{-10pt}
    \label{fig:ablation_steps}
\end{figure}

\begin{figure}[t]
    \centering
    \begin{subfigure}[b]{0.18\textwidth}
        \centering
        \includegraphics[width=\linewidth]{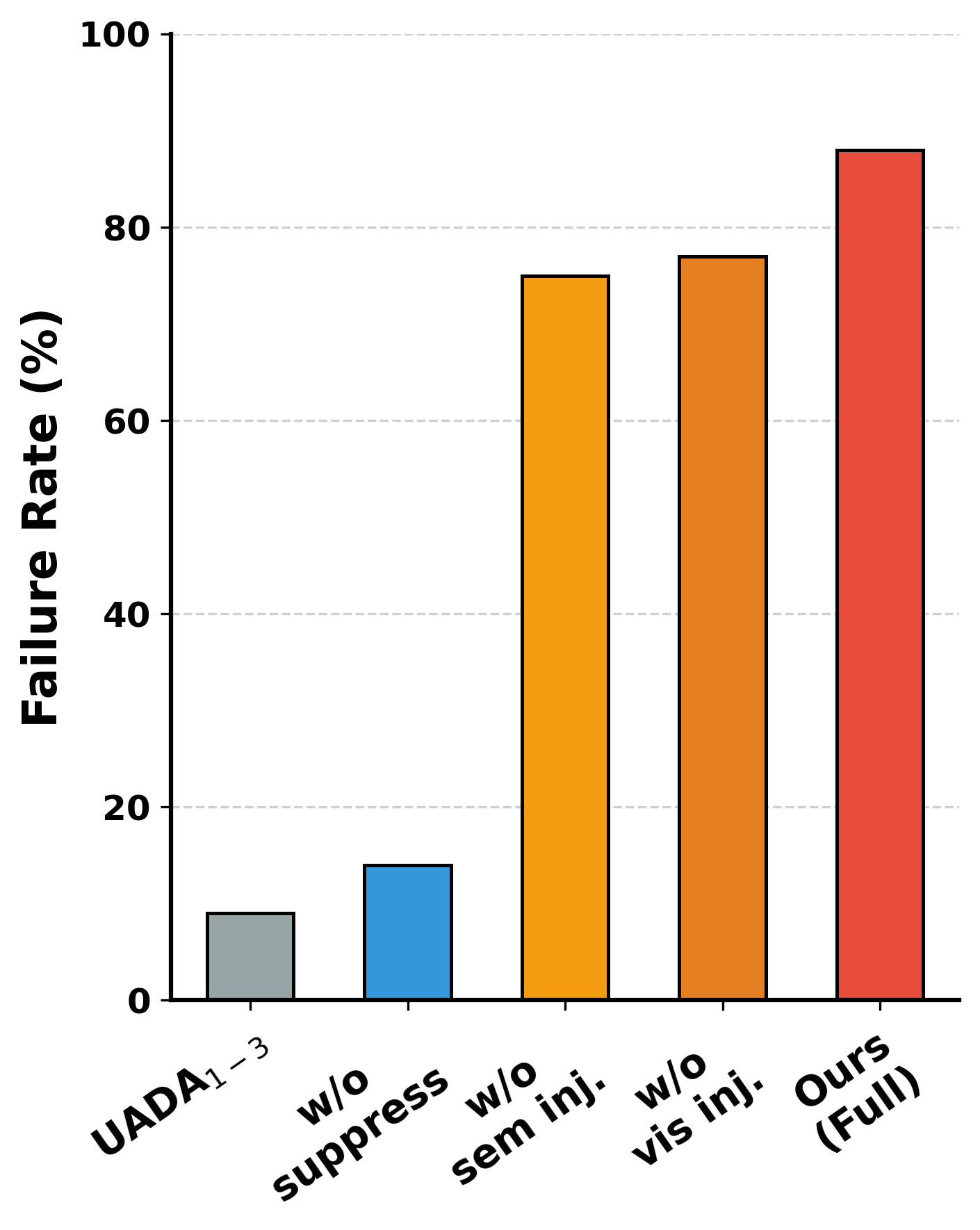} 
        \caption{UniVLA-Spa.}
    \end{subfigure}
    \hfill 
    \begin{subfigure}[b]{0.18\textwidth}
        \centering
        \includegraphics[width=\linewidth]{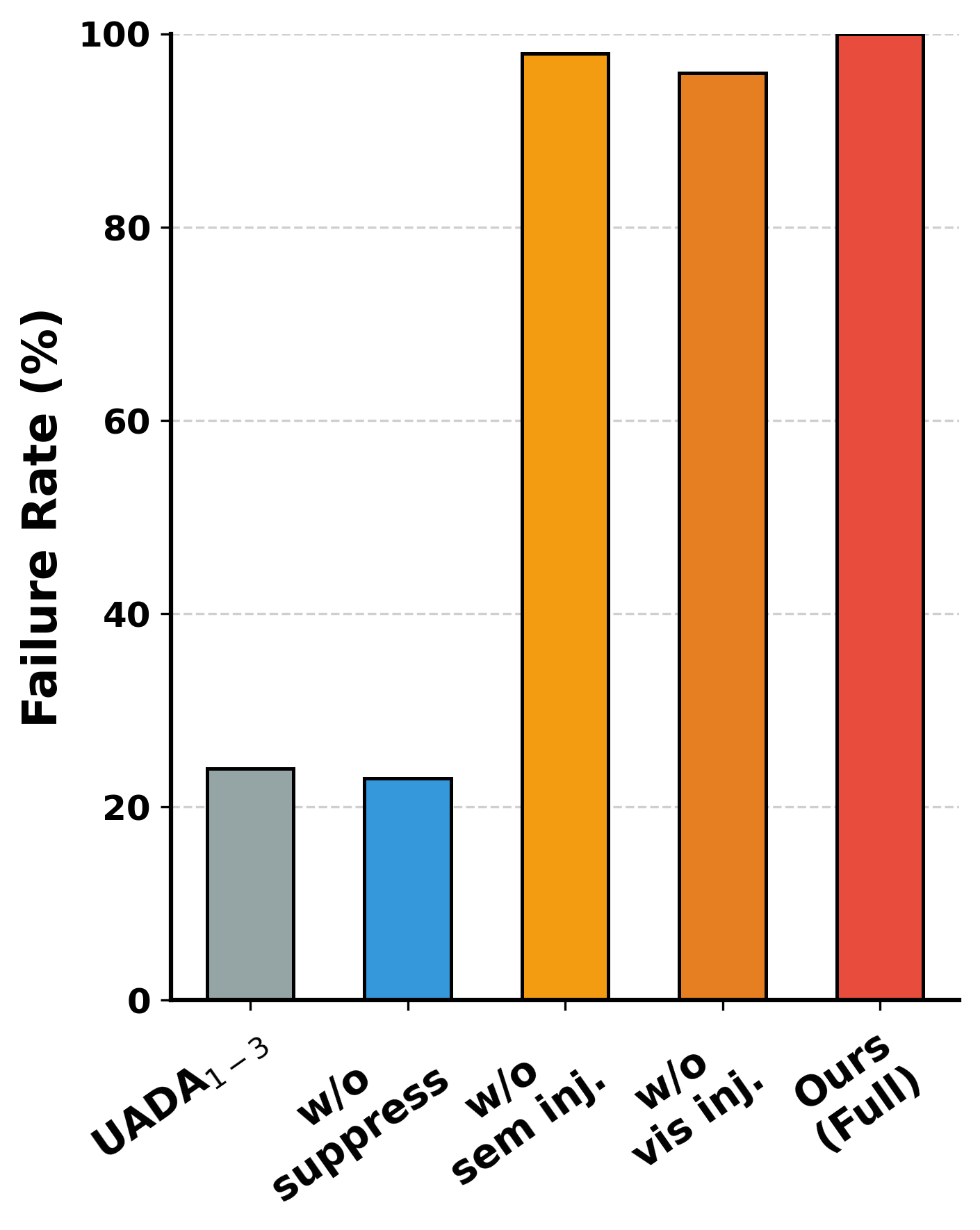}
        \caption{UniVLA-Obj.}
    \end{subfigure}
    \hfill 
    \begin{subfigure}[b]{0.18\textwidth}
        \centering
        \includegraphics[width=\linewidth]{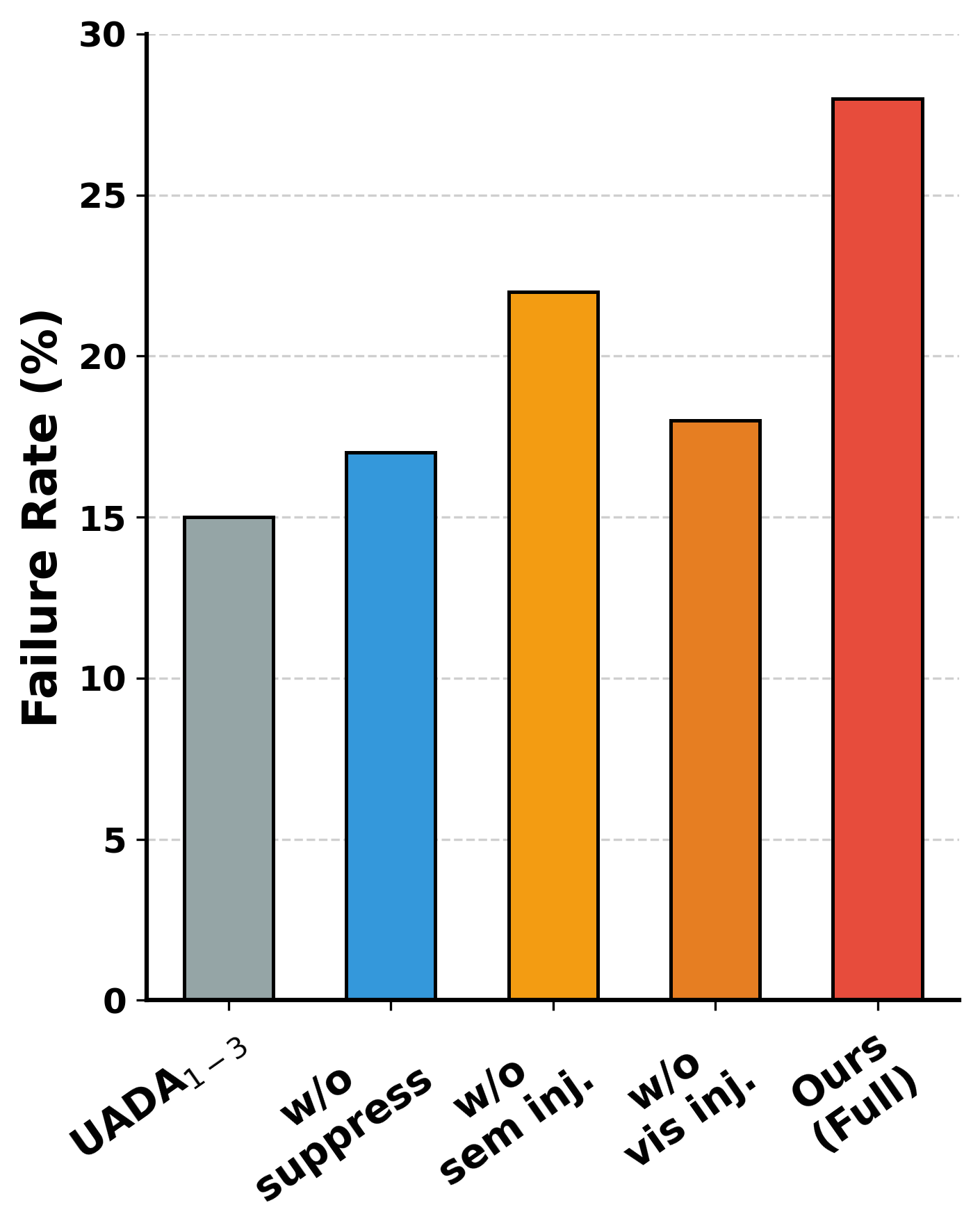}
        \caption{UniVLA-Goal}
    \end{subfigure}
    \hfill 
    \begin{subfigure}[b]{0.18\textwidth}
        \centering
        \includegraphics[width=\linewidth]{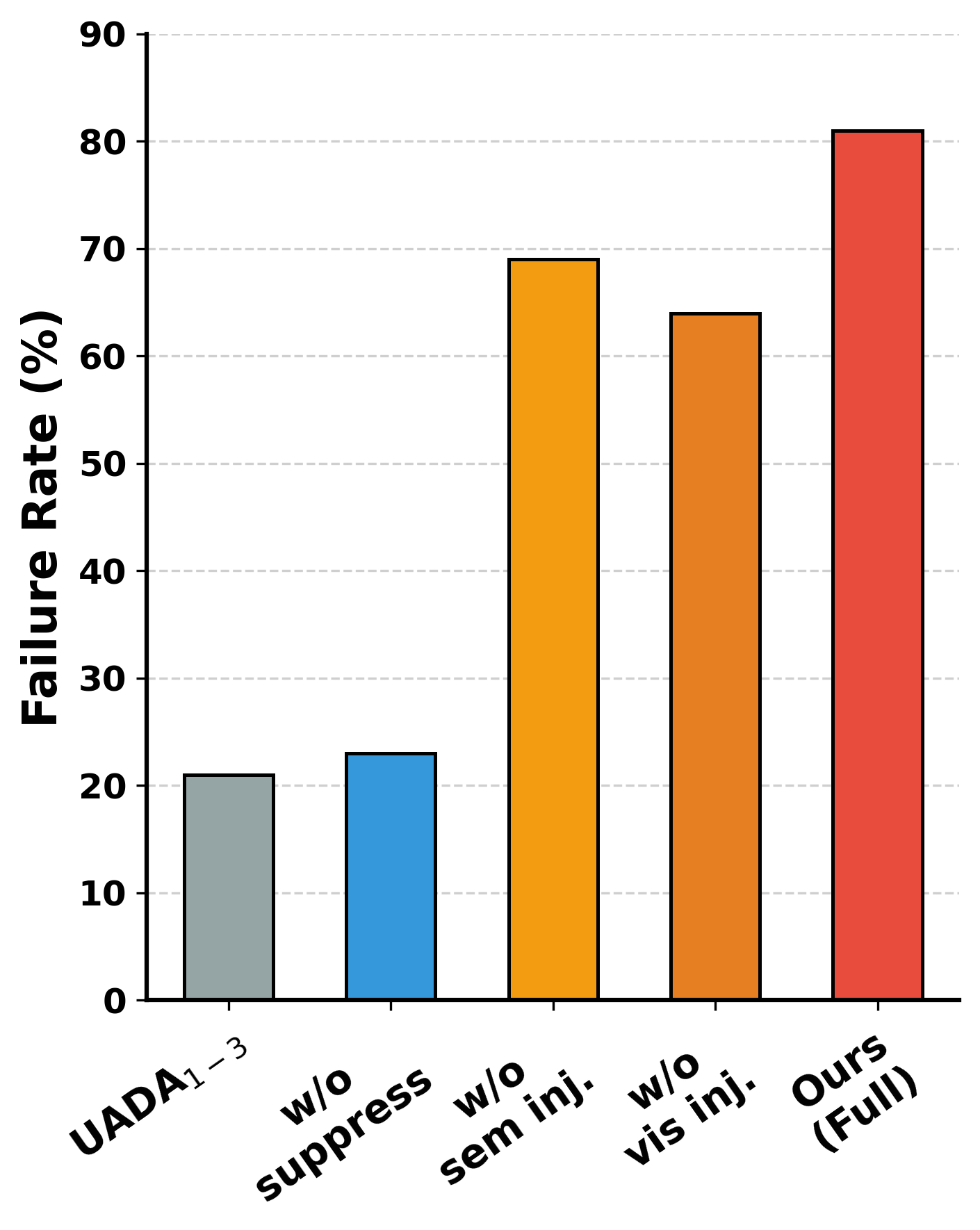}
        \caption{UniVLA-Lon.}
    \end{subfigure}
    \hfill 
    \begin{subfigure}[b]{0.18\textwidth}
        \centering
        \includegraphics[width=\linewidth]{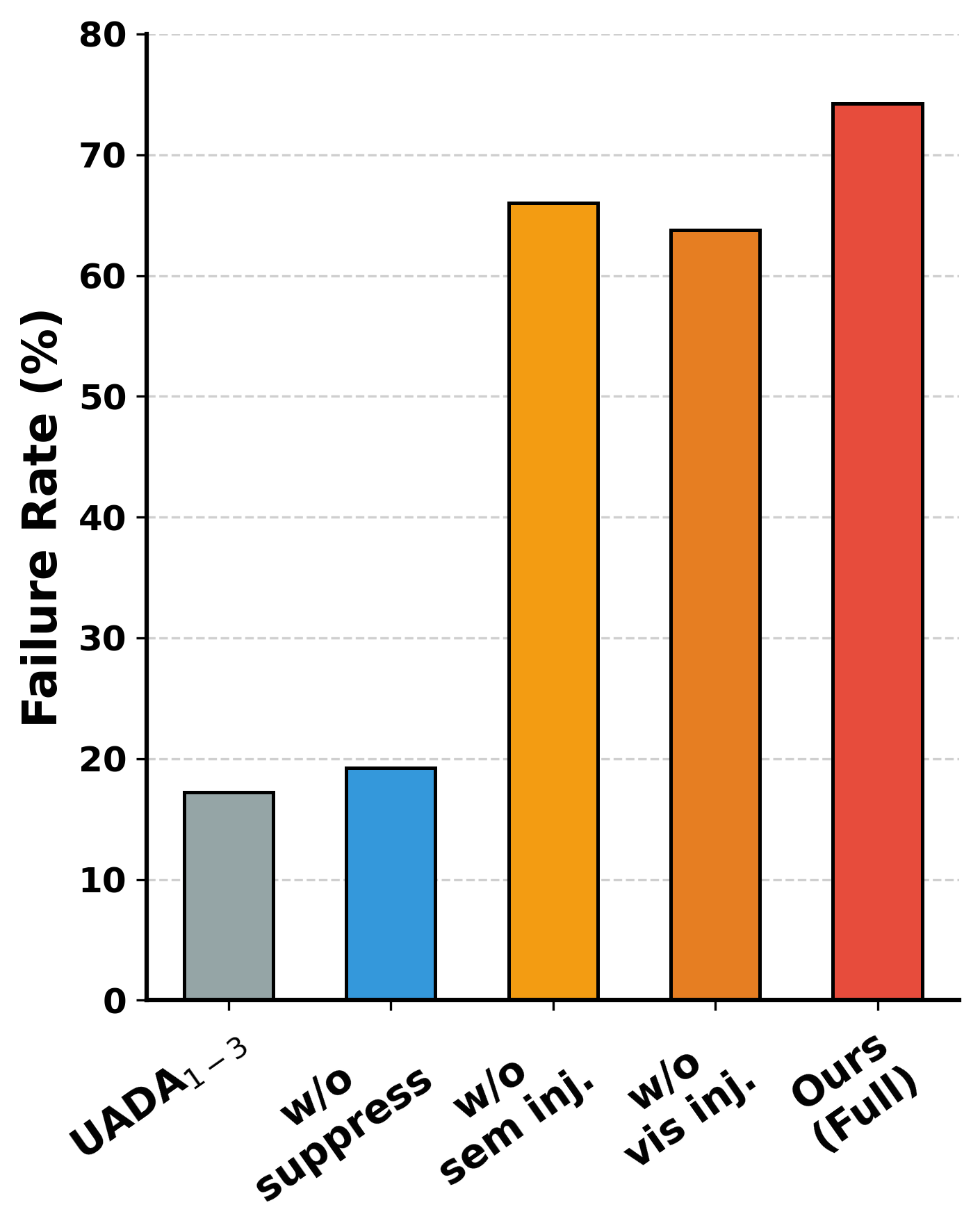}
        \caption{Average}
    \end{subfigure}
    
    \vspace{-6pt}
    \caption{\textbf{Ablation on Algorithmic Components.} We report the transfer Failure Rate (FR) of patches generated from \textit{OpenVLA-Spatial} and evaluated across four \textit{UniVLA} tasks (a-d), alongside their overall average (e).}
    \vspace{-5pt}
    \label{fig:component}
\end{figure}

\vspace{-10pt}

\subsubsection{Robustness across Surrogate Architectures.}
To verify VLA-Hijack's cross-architecture generalization, we evaluate distinct UniVLA variants as surrogates in Table~\ref{tab:3}. Baseline attacks (UADA, UPA, TMA)~\cite{uada} are inapplicable here, as their reliance on OpenVLA’s action discretization renders them incompatible with UniVLA’s architecture. This dependency renders them incompatible with UniVLA's architecture. In contrast, VLA-Hijack demonstrates superior compatibility across diverse architectures. Notably, the patch generated from UniVLA-Object achieves a Transfer Avg. of 64.35\%, demonstrating strong cross-architecture generalization to OpenVLA and CronusVLA. By targeting the shared Visual Proprioception Loop, our Proprioceptive Identity Rewiring mechanism ensures robust compatibility across heterogeneous VLA architectures.

\vspace{-6pt}

\subsection{Ablation Study}

\subsubsection{Impact of Attack Steps.}
To investigate the relationship between the number of attack steps and both convergence efficiency and transferability, we track the Failure Rate trajectories over 500 steps. The results (Fig.~\ref{fig:ablation_steps}) reveal two critical insights: 1) High White-box Optimization Efficiency. VLA-Hijack reaches 100\% FR within 30 steps by directly targeting the model’s decision logic with explicit perceptual objectives. Conversely, prior methods'~\cite{uada} indirect action-error objectives result in lower efficiency.
2) Superior Transfer Generalization. The contrast becomes stark in cross-architecture scenarios (Fig.~\ref{fig:ablation_steps}(c) and (d)). As optimization proceeds beyond 100 steps, baseline attacks~\cite{uada} (e.g., UADA$_{1-3}$, UPA$_1$) essentially flatline at a low FR. While baseline attacks flatline due to overfitting architecture-specific action tokens, VLA-Hijack achieves sustained transfer gains by synthesizing a ``phantom embodiment'' to progressively refine universal robotic features shared across architectures.

\vspace{-10pt}

\begin{table}[t]
    \centering
    \fontsize{6pt}{8pt}\selectfont
    
    \begin{minipage}{0.48\textwidth}
        \centering
        \caption{\textbf{Sensitivity to Injection Weight $\lambda$.} Transfer FR on UniVLA variants across different injection weights.}
        \label{tab:weight}
        \vspace{-6pt}
        \begin{tabular}{@{}c|ccccc@{}}
        \toprule
        \multirow{2}{*}{Weight $\lambda$~} & \multicolumn{5}{c}{\textbf{UniVLA Variants (Failure Rate \%)}}                                           \\ \cmidrule(l){2-6} 
                                   & Spatial        & Object          & Goal           & \multicolumn{1}{c|}{Long}           & \textbf{Avg}   \\ \midrule
        0.1                        & 79.00          & 98.00           & 18.00          & \multicolumn{1}{c|}{62.00}          & 64.25          \\
        \textbf{0.2}                        & \textbf{88.00} & \textbf{100.00} & \textbf{28.00} & \multicolumn{1}{c|}{\textbf{81.00}} & \textbf{74.25} \\
        0.3                        & 83.00          & 99.00           & 18.00          & \multicolumn{1}{c|}{80.00}          & 70.00          \\
        0.5                        & 75.00          & 99.00           & 18.00          & \multicolumn{1}{c|}{72.00}          & 66.00          \\ \bottomrule
        \end{tabular}
    \end{minipage}
    \hfill 
    \begin{minipage}{0.48\textwidth}
        \centering
        \caption{\textbf{Impact of Alternating Period $\tau$.} Comparing our alternating schedule against joint optimization (w/o alt.).}
        \label{tab:period}
        \vspace{-6pt}

        \begin{tabular}{@{}c|ccccc@{}}
        \toprule
        \multirow{2}{*}{Period $\tau$~} & \multicolumn{5}{c}{\textbf{UniVLA Variants (Failure Rate \%)}}                                                     \\ \cmidrule(l){2-6} 
                                       & Spatial        & Object          & Goal           & \multicolumn{1}{c|}{Long}           & \textbf{Avg}   \\ \midrule
        w/o alt.                            & 69.00          & 98.00           & 19.00          & \multicolumn{1}{c|}{79.00}          & 66.25          \\
        3                              & 74.00          & 98.00           & 20.00          & \multicolumn{1}{c|}{78.00}          & 67.50          \\
        \textbf{5}                              & \textbf{88.00} & \textbf{100.00} & \textbf{28.00} & \multicolumn{1}{c|}{\textbf{81.00}} & \textbf{74.25} \\
        10                             & 80.00          & \textbf{100.00} & 23.00          & \multicolumn{1}{c|}{85.00}          & 72.00          \\ \bottomrule
        \end{tabular}
    \end{minipage}
    \vspace{-12pt}

\end{table}

\subsubsection{Contribution of Individual Components.}
We ablate specific loss terms to quantify each module's impact (Fig.~\ref{fig:component}). Notably, removing the suppression term (w/o suppress) causes a catastrophic performance drop to near-baseline levels. This confirms our core motivation: without actively suppressing the original embodiment, the highly salient real arm dominates the model's attention, causing severe semantic interference that renders injected adversarial features ineffective. Furthermore, ablating either semantic anchoring (w/o sem inj.) or the visual projection (w/o vis inj.) leads to notable performance degradation. This indicates that while suppressing the real arm is a prerequisite, the patch still fails to establish a robust surrogate identity without both the high-level linguistic concept and the fine-grained textural realism of a "robotic arm." Ultimately, our full VLA-Hijack framework consistently achieves the highest Failure Rate across all tasks (Fig.~\ref{fig:component}e), demonstrating that creating a perceptual vacuum and filling it with a multimodal phantom embodiment are highly complementary and indispensable for robust cross-architecture transfer.

\vspace{-10pt}

\subsubsection{Optimization Strategy.}
Beyond individual components, we evaluate optimization strategies using patches generated on \textit{OpenVLA-Spatial} over 500 steps.

First, we analyze the necessity of our Alternating Injection Schedule across different periods $\tau$ (Table~\ref{tab:weight}). Replacing this schedule with naive joint optimization (\textbf{w/o alt.}) causes the average transfer FR to drop from \textbf{74.25\%} to \textbf{66.25\%}. This degradation occurs because simultaneously enforcing high-level semantic concepts and fine-grained visual prototypes forces the pixels into competing update directions, inducing severe gradient conflicts that prevent convergence to a unified surrogate identity. Decoupling these modalities across optimization steps resolves this interference, with $\tau=5$ providing the optimal update frequency.

Furthermore, we assess the sensitivity of the weighting factor $\lambda$ (Table~\ref{tab:period}), which regulates the trade-off between \textit{Proprioceptive Suppression} and \textit{Multimodal Injection}. As shown, the transfer performance exhibits a distinct peak at $\lambda=0.2$. When $\lambda$ is too small (e.g., 0.1), the injected features are insufficient to establish a robust phantom embodiment. Conversely, a larger $\lambda$ (e.g., 0.5) overwhelms the objective, degrading the suppression term and allowing the real arm's features to re-emerge and cause semantic interference. This confirms that VLA-Hijack relies on a carefully calibrated equilibrium between actively suppressing the original identity and injecting the new one.

\begin{figure}[t]
    \centering
    \begin{subfigure}[b]{0.15\textwidth}
        \centering
        \includegraphics[width=\linewidth]{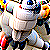} 
        \caption{\centering Spatial, \\OpenVLA-Sp.}
    \end{subfigure}
    \hfill 
    \begin{subfigure}[b]{0.15\textwidth}
        \centering
        \includegraphics[width=\linewidth]{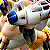}
        \caption{Bridge V2, \\ OpenVLA-7B}
    \end{subfigure}
    \hfill 
    \begin{subfigure}[b]{0.15\textwidth}
        \centering
        \includegraphics[width=\linewidth]{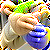}
        \caption{\centering Spatial, \\ UniVLA-Spa.}
    \end{subfigure}
    \hfill 
    \begin{subfigure}[b]{0.15\textwidth}
        \centering
        \includegraphics[width=\linewidth]{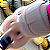}
        \caption{\centering Object, \\ UniVLA-Obj.}
    \end{subfigure}
    \hfill 
    \begin{subfigure}[b]{0.15\textwidth}
        \centering
        \includegraphics[width=\linewidth]{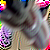}
        \caption{\centering Goal, \\ UniVLA-Goal}
    \end{subfigure}
    \hfill 
    \begin{subfigure}[b]{0.15\textwidth}
        \centering
        \includegraphics[width=\linewidth]{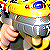}
        \caption{\centering Long, \\ UniVLA-Long}
    \end{subfigure}
    \vspace{-3pt}

    \caption{\textbf{Visualizations of Adversarial Patches.} Patches optimized on various surrogate models consistently manifest distinct structural features of physical robotic arms.}
    \label{fig:vis_analy}
    \vspace{-10pt}
\end{figure}

\vspace{-3pt}

\subsection{Patch Pattern Analysis}
To qualitatively validate the efficacy of our \textit{Proprioceptive Identity Rewiring} mechanism, we visualize the optimized adversarial patches generated from various surrogate models in \textbf{Fig.~\ref{fig:vis_analy}}. As observed across all subfigures, whether optimized on simulation tasks (LIBERO)~\cite{libero} or real-world physical datasets (BridgeData V2)~\cite{bridgedata}, the generated patches consistently exhibit highly structured visual characteristics. Specifically, they strikingly manifest the physical components of robotic arms, displaying clear articulated joint structures, metallic textures, and the geometric shapes of mechanical grippers. Such structural mimicry visually confirms that VLA-Hijack successfully synthesizes a "phantom embodiment" to deceive the model's proprioceptive loop. Detailed visualizations of the hijacked execution rollouts are provided in the Supplementary Materials.

\vspace{-4pt}

\section{Conclusion}
In this work, we identify a fundamental vulnerability in VLA models: their critical reliance on a visual proprioception loop. To break the transferability bottleneck of existing action-space attacks, we propose VLA-Hijack, a novel adversarial framework that severs the semantic relationship between the agent's physical embodiment and its control policy. By synergistically coupling attention-guided proprioceptive suppression with alternating multimodal injection, VLA-Hijack synthesizes a "phantom embodiment" to hijack the model's decision-making. Extensive experiments demonstrate that our method achieves rapid white-box convergence and sets a new state-of-the-art for black-box transferability across heterogeneous architectures and cross-domain environments. 
\clearpage  


%
%
\bibliographystyle{splncs04}
\bibliography{main}
\end{document}